\documentclass[10pt,twocolumn,letterpaper]{article}

\usepackage{cvpr}
\usepackage{times}
\usepackage{epsfig}
\usepackage{graphicx}
\usepackage{amsmath}
\usepackage{amssymb}
\usepackage{float}
\usepackage{subcaption}
\usepackage{slashbox}
\usepackage{bbm}
\usepackage{enumitem}
\usepackage{dblfloatfix}

\newcommand{\mb}{\mathbf}
\newcommand{\tb}{\textbf}

\DeclareMathOperator*{\argmin}{arg\,min}

\usepackage[pagebackref=true,breaklinks=true,letterpaper=true,colorlinks,bookmarks=false]{hyperref}

\cvprfinalcopy 

\def\cvprPaperID{8306} 

\ifcvprfinal\fi

\graphicspath{{figs/}}

\begin{document}
\setlength{\abovedisplayskip}{0.08in}
\setlength{\belowdisplayskip}{0.08in}

\title{Physically Realizable Adversarial Examples for LiDAR Object Detection}

\author{James Tu${}^{1}$ \quad  Mengye Ren${}^{1,2}$ \quad  Siva Manivasagam${}^{1,2}$ \quad  Ming Liang${}^{1}$\\
Bin Yang${}^{1,2}$ \quad  Richard Du${}^3$ \quad  Frank Cheng${}^{1,2}$ \quad  Raquel Urtasun${}^{1,2}$\\
{\tt\small \{james.tu,mren3,manivasagam,ming.liang,byang10,frank.cheng,urtasun\}@uber.com}\\{\tt\small rdu@princeton.edu}\\
${}^1$Uber ATG \quad ${}^2$University of Toronto \quad ${}^3$Princeton University
}

\maketitle

\begin{abstract}
Modern autonomous driving systems rely heavily on deep learning models to process point cloud
sensory data; meanwhile, deep models have been shown to be susceptible to adversarial attacks with
visually imperceptible perturbations. Despite the fact that this poses a security concern for the self-driving
industry,  there has been very little  exploration  in terms of 3D
perception, as most adversarial attacks have only been applied to 2D flat images. In this paper, we address this issue and present a
method to generate universal 3D adversarial objects to fool LiDAR detectors. In particular, we
demonstrate that placing an adversarial object on the rooftop 
of any target vehicle to hide the vehicle entirely from LiDAR detectors with a success rate of 80\%. We report attack
results on a suite of detectors using various input representation of point clouds.  We also  conduct a pilot study on adversarial defense using data augmentation. This is one step
closer towards safer self-driving under unseen conditions from limited training data.

\end{abstract}

\vspace{-0.2in}

\section{Introduction}
\vspace{-0.05in}

Modern autonomous driving systems use deep neural networks (DNNs) to process LiDAR point clouds
in order to perceive the world \cite{mmf,pointrcnn,voxelnet}. Despite introducing significant performance improvements, DNNs have been previously found to be vulnerable to adversarial attacks when using
image inputs~\cite{szegedy2014intriguing,goodfellow2015explaining,kurakin2017physical,athalye2018synthesizing},
where a small perturbation in the input pixels can cause drastic changes in the output predictions.
The potential vulnerabilities, in conjunction with the safety-critical nature of self-driving,
motivate us to investigate the possibility of disrupting autonomous driving systems with
adversarial attacks.


Image perturbations alone however, are not enough for modern autonomous driving systems, which are
typically equipped with LiDAR sensors producing point clouds as the primary main sensory input.
Several previous works have shown successful
attacks~\cite{Wicker_2019_CVPR,Xiang_2019_CVPR,Zeng_2019_CVPR} with point cloud perturbations,
generating salient modifications by adding, removing, and modifying points. Although these attacks
work in theory, arbitrary to point clouds are not always physically realizable. For
instance, a given point cloud may be impossible to generate from a LiDAR sensor due to the lasers' 
fixed angular frequencies and light projection geometry. 

Towards generating physically realizable attacks, Cao \etal \cite{cao2019adversarial} propose to
learn an adversarial mesh capable of generating adversarial point clouds with a LiDAR renderer.
However, their work only considers learning an adversarial mesh for a few specific frames. As a result, the
learned 3D object is not \textit{universal} and may not be reused in other 3D scenes. Moreover, they have only evaluated their
attack on a very small in-house dataset that contains around a few hundred frames.

\begin{figure}[t]
\includegraphics[width=1.\linewidth]{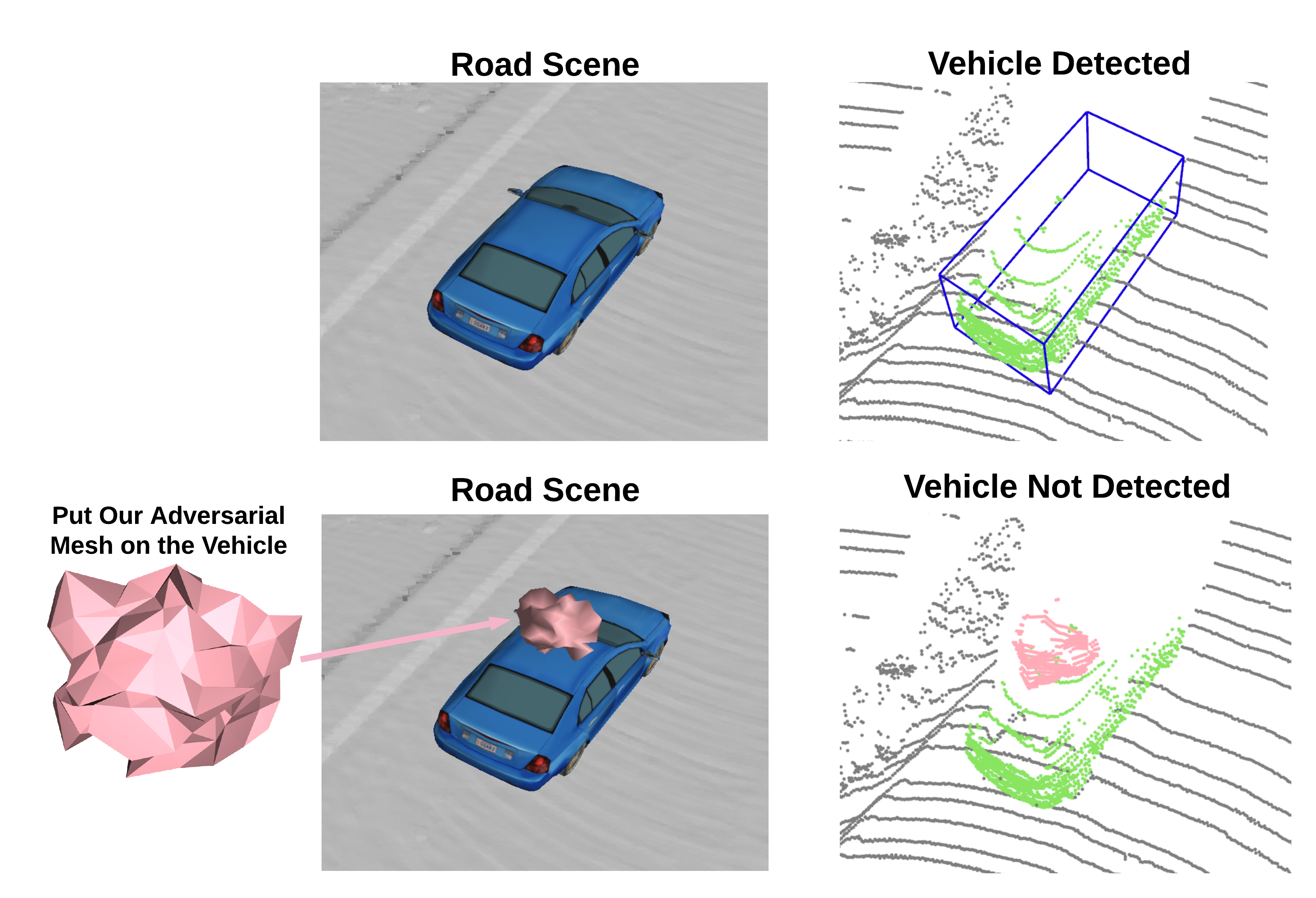}
\vspace{-0.25in}
\caption{In this work we produce a physically realizable adversarial object that can make vehicles
``invisible''. After placing the object on the rooftop of a target vehicle, the vehicle will no
longer be detected by a LiDAR detector.}
\vspace{-0.15in}
\label{fig:teaser}
\end{figure}

In contrast to \cite{cao2019adversarial}, we aim to learn a universal and physically realizable
adversary. Furthermore, we craft 3D objects in a novel setting where they are placed on top of
vehicles as rooftop cargo.
Such objects can be used in any scene and on any
types of small vehicles, and can hide the entire host vehicle from a strong LiDAR detector
\cite{pixor} with a success rate of 80\% at IoU 0.7. By comparison, placing a random object on the
rooftop only produces a small drop in detection accuracy. We evaluated our learned adversarial
object on a suite of common LiDAR detector architectures that take in various kinds of input
representations, and we also report transferability of attacks across these models. Lastly, we
conduct a pilot study on adversarial defense using data augmentation and adversarial training. Applying defense mechanisms significantly decreases the likelihood of missing detections of vehicles with strange roof-top cargo, which is a rare-seen but practical scenario for self-driving cars.

The contributions of this paper can be summarized as follows:
\vspace{-0.05in}
\begin{enumerate}[noitemsep]
    \item We propose a universal adversarial attack on LiDAR detectors with physically realizable 3D meshes.
    \item We present a novel setting where the adversarial object makes the target vehicle invisible
    when placed on the vehicle rooftop.
    \item We report a thorough evaluation across different detector, each using different input representations. 
    \item We present a successful defense mechanism via training with data augmentation.
\end{enumerate}
\vspace{-0.05in}

In the following, we first review prior literature on adversarial attacks and in particular point
cloud and 3D physical attacks. Next we present details of our proposed method to generate a
physically realizable adversarial object, followed by empirical evaluations of our attack and defense on several LiDAR object detectors.




\vspace{-0.05in}
\section{Related Work}
\vspace{-0.05in}
Despite the impressive performance of deep learning models, they are surprisingly vulnerable to
minuscule perturbations. Adversarial attacks add visually imperceptible noise to the input to 
drastically alters a neural network's output and produce false predictions.


\vspace{-0.1in}
\paragraph{Image attacks:}
Adversarial examples were first discovered in the context of image classification
networks~\cite{szegedy2014intriguing}. These vulnerabilities were later discovered in networks
performing image detection and semantic segmentation as well~\cite{xie2017adversarialdetect}.
Attacks can either be \textit{white box}~\cite{goodfellow2015explaining,deepfool,pgd}, where the target model's weights are available,  or \textit{black box} \cite{decision,transfer,zoo}, where
the adversary can only query the outputs of the target model. Various defense mechanisms, including
adversarial training~\cite{goodfellow2015explaining,pgd}, denoiser~\cite{denoiser}, Bayes
classifier~\cite{bayesclassifier}, certified defense~\cite{provable,certified} have been proposed,
and shown effective on a certain range of attack types.

\vspace{-0.1in}
\paragraph{Point cloud attacks:} With the rise of  LiDAR sensors in robotics applications  such as  self-driving, point cloud data has become a popular input representation.  Some recent
research demonstrated the possibility of adversarial attacks on networks that take  point cloud
data as input. \cite{yang2019adversarial} proposed to add, remove, or perturb points; whereas
\cite{Xiang_2019_CVPR} added clusters of adversarial points.
\cite{zheng2018learning,Wicker_2019_CVPR} proposed saliency-based approaches for removing points.
Several structured methods have been also introduced to perturb point clouds with the goal of preserving
physical fidelity~\cite{liu2019adversarial}.

\vspace{-0.12in}
\paragraph{Physical world attacks:}
Perturbing image pixels or point locations alone may not guarantee that the attack can happen in the
physical world. To address this issue, several works have produced physical adversaries and expose
real world threats. \cite{kurakin2017physical} studies whether the image pixel perturbations can be
realized by a physical printer. \cite{advpatch} produces a universal and robust adversarial sticker.
When placed on any image with any pose, the sticker induces a targeted false classification.
\cite{eykholt2018robust} proposes to train the attack with different view angles and distances to
make it robust. \cite{athalye2018synthesizing} synthesizes robust adversarial 3D objects capable of
fooling image classifiers when rendered into an image from any angle.
\cite{Zeng_2019_CVPR,liu2019diffrenderer} consider perturbing other photo-realistic properties such
as shape normal and lighting, using a differentiable renderer. Nevertheless, in these approaches,
the final outputs are still projected to 2D image space.

In the context of self-driving and LiDAR perception, \cite{cao2019adversarial} propose
\textit{LidarAdv}, a method to learn adversarial meshes to fool LiDAR detectors. Our work is
different from theirs in several important ways. First, \textit{LidarAdv} only considers one frame during
learning and hence is input specific; whereas we train our adversary on all frames and all vehicles,
creating a universal adversary. Second, our adversary can be placed on a vehicle roof to hide it, whereas their adversarial object does not interact with other real world
objects. Lastly, we are the first to conduct a thorough evaluation of a physically realizable adversarial attack on a
suite of detectors and on a public large scale LiDAR benchmark. Besides learning a physical mesh,
\cite{cao2019sensor} proposes to use laser devices to spoof LiDAR points. 
These laser devices,
however, are more difficult to set up and it is not trivial to create consistent point clouds as the sensor
moves.

\section{Physically Realizable Adversarial Examples}
\vspace{-0.05in}
\begin{figure*}
\vspace{-0.2in}
\centering
\includegraphics[width=1.\linewidth]{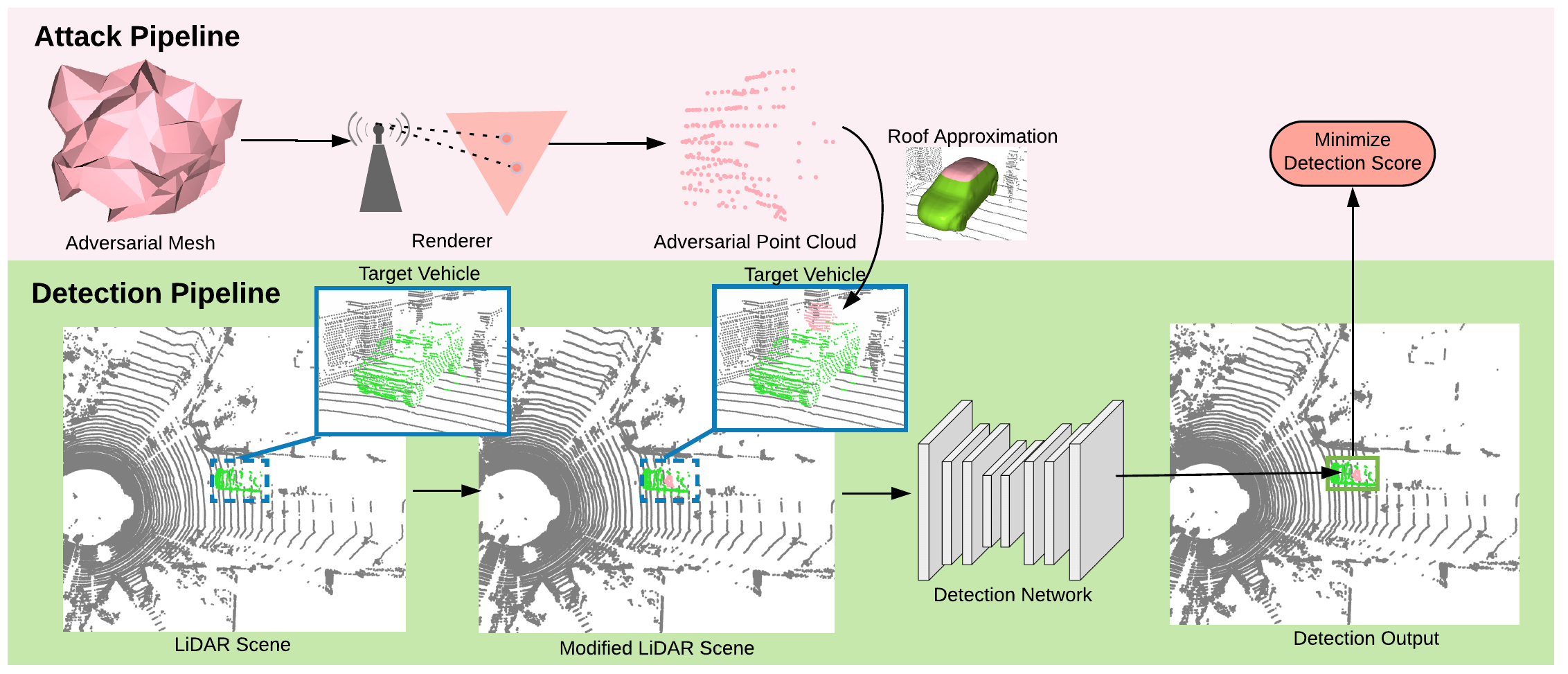}
\vspace{-0.22in}
\caption{Overall adversarial example generation pipeline. We start from a mesh representation, and
use a LiDAR renderer to obtain the point clouds. By using roof approximation techniques, we attach
the adversarial point cloud on top of the target vehicle, and we modify the mesh vertices so that
the detection confidence score of the target vehicle is minimized. 
}
\label{fig:pipeline}
\vspace{-0.12in}
\end{figure*}

In this section, we present our method for learning an adversarial object to attack LiDAR detectors.
There are many possible ways to place an object in a scene for an adversarial effect, as explored in
\cite{cao2019adversarial}. For example, one can hide the inserted adversarial object or change the
object label (\eg, making the detector believe that the adversarial object is a pedestrian). In this
paper, we instead focus on a novel setting, where we place the object on the rooftop of a vehicle
and hide the vehicle from the detector, hence creating an ``invisible'' car, illustrated in
Figure~\ref{fig:pipeline}. Such a procedure can be easily reproduced in the real world and is also
plausible even without the presence of a malicious attacker, since cars occasionally carry pieces of
furniture or sports equipment on their rooftops.

For the rest of this section, we first describe the 3D representation of the adversarial example and
how to render it into a point clouds. We then present  our adversarial example generation algorithms.
Lastly, a simple defense algorithm based on data augmentation is presented.

\vspace{-0.05in}
\subsection{Surface Parameterization}
\vspace{-0.05in}
Many parameterizations exist for 3D objects, including voxels, meshes, and implicit
surfaces~\cite{foley1996computer}. Voxels are easy to compute but require significantly more memory
than the alternatives to produce a high level of detail. Implicit surfaces, on the other hand,
provide compact representations but are harder to render since they require solving for the
numerical roots of the implicit functions. In this paper, we choose to represent our adversary with
a mesh since it benefits from compact representations and allows for efficient and precise
rendering. Given a mesh, we can compute the exact intersections of rays analytically and in a
differentiable manner. The latter is important since it allows us to take gradients efficiently for
white box attacks. Furthermore, meshes have previously demonstrated high-fidelity shape generation
results on faces and human bodies~\cite{Bagautdinov_2018_CVPR,Litany_2018_CVPR}.

During the learning of the adversarial mesh, following prior
literature~\cite{liu2019soft,kato2018neural}, we deform a template mesh by adding local learnable
displacement vectors $\Delta \mb{v}_i \in \mathbb{R}^{3}$ for each vertex and a global
transformation for the entire mesh,
\begin{align}
\mb{v}_i = \mb{R} (\mb{v}_i^0 + \Delta \mb{v}_{i}) + \mb{t},
\end{align}
where $\mb{v}_i^0$ is the initial vertex position, and $\mb{R} \in SO(3)$ is a global rotation
matrix, and $\mb{t} \in \mathbb{R}^{3}$ is a global translation vector. To ensure physical
feasibility, box constraints are applied to the mesh vertices as well as the global translation.

In the experiments where we initialize the mesh from an isotropic sphere, $\mb{R}$ is fixed to be
the identity matrix, since the sphere is rotation invariant. In the experiments where we deform
common objects, we constrain $\mb{R}$ to be rotations on the $x$-$y$ plane:
\begin{align}
R=\begin{bmatrix}
\cos\theta & -\sin\theta & 0 \\
\sin\theta & \cos\theta & 0 \\
0 & 0 & 1
\end{bmatrix},
\end{align}
where $\theta$ is the learnable rotation angle.

\begin{figure*}
\vspace{-0.25in}
\centering
\includegraphics[trim=0cm 0.5cm 0cm 0cm, clip, width=1.\linewidth]{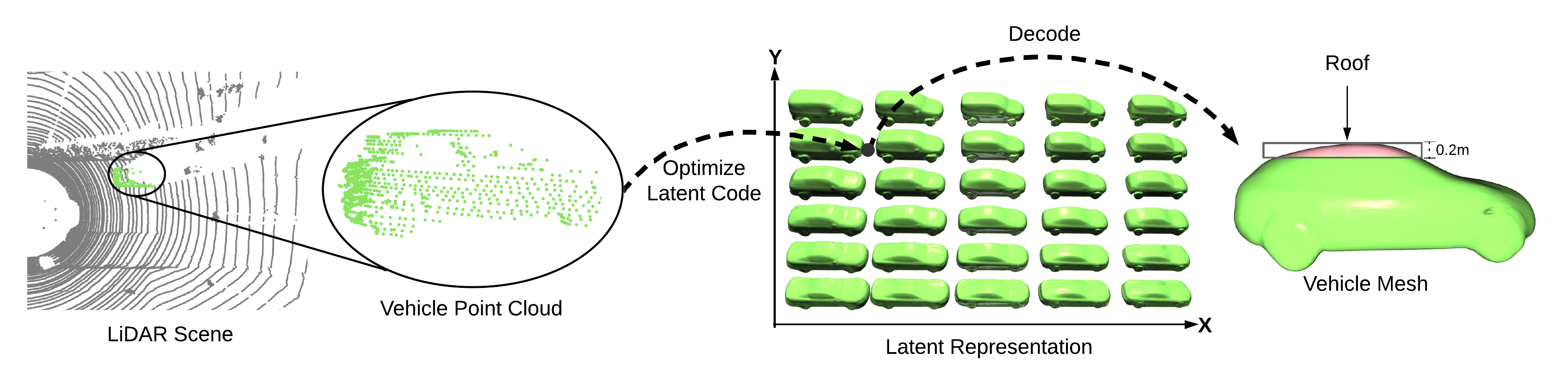}
\vspace{-0.2in}
\caption{Approximating rooftop from vehicle point clouds. We build a low dimensional
representation of our vehicle object bank using PCA, and embed the target vehicle point clouds by
optimizing the latent code. The top 0.2m is then cropped to be the rooftop region.}
\label{fig:mesh_fit}
\vspace{-0.2in}
\end{figure*}

\vspace{-0.05in}
\subsection{LiDAR Simulation}
\vspace{-0.05in}
We aim to add an adversarial mesh $Z$ into the scene in a realistic manner and choose the roof of
vehicles as the location for placement, as placing objects on top is easier due to gravity and does
not interfere with adjacent traffic.  Furthermore, objects on top of vehicles are less prone to
occlusion, whereas areas like the front hood or trunk top may be blocked by another vehicle. Finally, this is a realistic scenario as it common to strap furniture,
canoes, bicycles, and other large items on top of vehicles. In this section we first describe how to
render a mesh into LiDAR points.  Next we introduce a technique to locate the
rooftop region from a vehicle point cloud, where we can place the adversary.

\vspace{-0.15in}
\paragraph{LiDAR point rendering:} We then use location of the mesh in the scene to sample nearby
rays $\gamma$ with the same angular frequencies as the LiDAR sensor used to generate the original
LiDAR scene. Given rays $\gamma$ and mesh $Z$, the adversarial points  are rendered with a
differentiable raycaster $R$. We compute the intersection of rays and mesh faces with the
Moller-Trumbore intersection algorithm~\cite{moller}. We refer readers to the supplementary
materials for more details on this. Then, we take the union of the rendered adversarial
points and the original points to create the modified scene.

\vspace{-0.1in}
\paragraph{Rooftop fitting:}
To approximate the center of a vehicle's roof from its point cloud, as illustrated in
Figure~\ref{fig:mesh_fit}, we first fit a CAD model to the point cloud. Inspired by \cite{sampl}, we
represent our internal collection of vehicle models as signed distance functions (SDFs), denoting as
$F(\cdot;\theta)$, and project this library of vehicles into a latent space using PCA. Since SDFs
implicitly represent 3D surfaces as its zero level-set, we optimize the latent code $z$ such that
all ground truth vehicle points evaluate as close to 0 as possible. Concretely, given a vehicle
bounding box $(x, y, w, h, \alpha)$, and a set of points $P = \{p : p \in \mathbb{R}^{3}\}$ within
the box, we find the optimal latent code $z^{*}$ such that
\begin{align}
    z^{*} = \argmin_{z} \sum_{p \in P} {F(p ; \theta(z))}^{2}.
\end{align}
We then apply marching cubes~\cite{foley1996computer} on $F(\cdot ; \theta(z^{*}))$ to obtain a
fitted CAD model. Lastly, we use vertices within the top 0.2m vertical range of the CAD model to
approximate the roof region.

During the attack, we place the adversarial mesh $Z$ with a fixed pose relative to the roof center
of a target vehicle. Given a vehicle bounding box $(x, y, w, h, \alpha)$, we compute the roof center
$(r_x, r_y, r_z)$ and apply transformation matrix
\begin{align}
T=\begin{bmatrix}
\cos\alpha & -\sin\alpha & 0 & r_x \\
\sin\alpha & \cos\alpha & 0 & r_y \\
0 & 0 & 1 & r_z \\
0 & 0 & 0 & 1
\end{bmatrix}
\end{align}
on the adversarial object.

\subsection{Adversarial Example Generation}
\vspace{-0.05in}
In this section, we first introduce the objective function being optimized to learn adversarial examples. Both white box and black box attack algorithms are presented next.

\vspace{-0.05in}
\subsubsection{Objective}
\vspace{-0.05in}
The overall loss function is a combination of the adversarial loss and the Laplacian loss for mesh smoothness:
\begin{align}
\mathcal{L} = \mathcal{L}_{\rm adv} + \lambda \mathcal{L}_{\rm lap}.
\end{align}
 To generate an adversarial example, we search for vertex perturbation $\mb{v}$ and global
transformation parameters $(\mb{R}, \mb{t})$ that minimize the loss function.

For the adversarial loss, following prior work~\cite{xie2017adversarialdetect}, we also find it
necessary to suppress all relevant bounding box proposals. A proposal is relevant if 1) its
confidence score is greater than 0.1 and 2) if its IoU with the ground truth bounding box is also
greater than 0.1.

Our adversarial objective minimizes the confidence of the relevant candidates:
\begin{align}
\mathcal{L}_{\rm adv} = \sum_{\substack{y,s \in \mathcal{Y}}} - \mathrm{IoU}(y^*, y) \log(1 - s),
\end{align}
where $\mathcal{Y}$ is the set of relevant bounding box proposals and each proposal $y$ has a
confidence score $s$. We use binary cross entropy to minimize the confidence score of the relevant
proposals, weighed by the IoU with the ground truth bounding box $y^*$. Here, we choose to target negative classification labels instead of other bounding box parameters because missing detections are the most problematic.

In addition, a Laplacian loss~\cite{liu2019soft} is applied to regularize mesh geometry and maintain
surface smoothness:
\begin{align}
\mathcal{L}_{\rm lap} = \sum_{i} {\lVert {\delta}_i \rVert}^2_2,
\end{align}
where $\delta_i$ is the distance from $v_i$ to the centroid of its immediate neighbors $N(i)$:
\begin{align}
\delta_i = v_i - \frac{1}{\lVert N(i) \rVert} \sum_{j \in N(i)} v_j.
\end{align}

\subsubsection{Attack Algorithms}
\vspace{-0.1in}
In this section we provide details for both white box and black box attacks for learning mesh
vertices.
\vspace{-0.1in}
\paragraph{White box attack:} 
In a white box setting, we simulate the addition of the adversary in a differentiable manner, and
hence can take the gradient from the objective $\mathcal{L}$ to the mesh vertices. In
addition, we re-parameterize local and global displacement vectors to apply box constraints, as
\cite{carlini2017towards} have demonstrated issues with other alternatives. Specifically, since
clipping parameters during \textit{projected gradient descent} creates a disparity between parameter
updates and momentum updates, we instead re-parameterize mesh vertices to inherently obey box
constraints:
\begin{align*}
\mb{v}_i = \mb{R}(\mb{b} \odot \mathrm{sign}(\tilde{\mb{v}}_i^0) \odot \sigma(\lvert \tilde{\mb{v}}_i^0 \rvert + \Delta \tilde{\mb{v}}_{i})) + \mb{c} \odot \tanh(\tilde{\mb{t}}),
\end{align*}
where $\odot$ denotes element-wise multiplication, $\sigma$ denotes the sigmoid function, $\mb{b}\in
\mathbb{R}^3$ define limits on size, and $\mb{c}\in \mathbb{R}^3$ define limits on translation.
$\sigma(\tilde{\mb{v}}_i^0) = \mb{v}_i^0 / \mb{b}$ is the normalized initial position of vertex and
$\tanh(\tilde{t}) = \mb{t} / \mb{c}$ is the normalized global translation. The $\mathrm{sign}$
function constrains each vertex to stay in its initial quadrant.

\vspace{-0.1in}
\paragraph{Black box attack:} A gradient-based attack is not always feasible in point cloud
perception due to non-differentiable preprocessing stages that are common in modern point cloud
detection models \cite{pixor, pointpillar}. For example, models like PIXOR \cite{pixor} represent
the input as occupancy voxels, preventing gradients from reaching the point cloud. To address this
problem, we employ a genetic algorithm \cite{alzantot2019genattack} to update the mesh parameters. Here, a population of candidates meshes are evolved to maximize the fitness score $-\mathcal{L}$. At every iteration, the candidate with the highest fitness is preserved while the rest are replaced. New candidates are generated by sampling mesh parameters from a pair of old candidates, with sampling probability proportional to fitness score. We then add gaussian noise to some new candidates sampled with a mutation probability. 
To jointly optimize over all samples, we perform inference on multiple examples and take the average fitness score at each iteration.
In this black box setting, we find re-parameterization unnecessary for gradient-free optimization.

\subsection{Defense Mechanisms}
Given that rooftop objects are rarely observed in the training distribution and that our attack
produces examples that are heavily out-of-distribution, we first propose random data
augmentation as a simple defense mechanism.  
Next, we consider adversarial training~\cite{carlini2017towards} for a stronger defense against adversarial attacks.
\vspace{-0.1in}
\paragraph{Data augmentation:}
When training with data augmentation, in every frame we generate a random
watertight mesh and place it on a random vehicle using the  methods presented
previously. This method is not specific to the type of optimization employed by the
attacker (\eg white box or black box) and hence may generalize better when compared to regular
adversarial training~\cite{goodfellow2015explaining}.
 
To generate a random watertight mesh, we first sample a set of $N$ vertices $V \in \mathbb{R}^{N
\times 3}$ from a Gaussian $\mathcal{N}(0, \sigma)$ and apply incremental triangulation to obtain a
set of connected tetrahedrons $Q$. We then stochastically remove $M$ boundary tetrahedrons that do
not disconnect $Q$ into separate components. Finally, we take the remaining boundary faces of $Q$ to
obtain a watertight surface. 
\vspace{-0.1in}
\paragraph{Adversarial training:}
While adversarial training has empirically been found to be robust, it is expensive and infeasible when the cost of training an adversary is high. Thus, we employ a method similar to~\cite{Wong2020Fast} and take one mesh update step per model update instead of a full optimization cycle. During training, the adversary is randomly re-initialized every $k$ steps using the same mesh generation method.



\vspace{-0.05in}
\section{Experiments}
\vspace{-0.05in}
In this section, we first discuss the datasets and models used in our experiments in
Section~\ref{sec:datasets} and \ref{sec:targets}, and the experimental setup in
Section~\ref{sec:setup}. We then present experimental results on 1) white box and black box attacks,
2) attacks and transferability on various detection backbones, 3) common object attacks, and 4)
adversarial defense using data augmentation training.


\begin{figure*}
\vspace{-0.15in}
\centering
\includegraphics[width=1\textwidth]{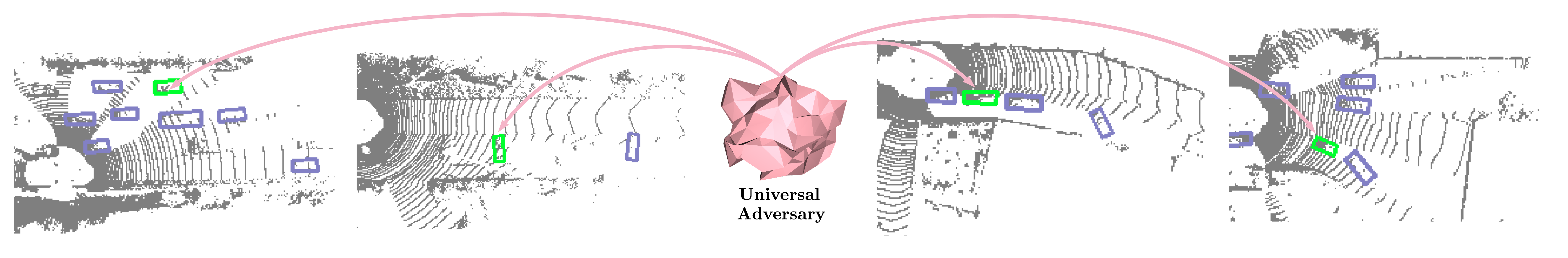}
\vspace{-0.35in}
\caption{Visualization of our universal adversarial object hiding different car instances at various orientations and locations.}
\vspace{-0.05in}
\label{fig:qualitative}
\end{figure*}

\begin{table*}[t]
\begin{minipage}[b]{0.37\textwidth}
    \centering
    \resizebox{\textwidth}{!}{
    \begin{tabular}{c | c | c}
    \backslashbox{Source}{Target} & PIXOR-1 & PIXOR-2 \\ \hline
    PIXOR-1 & \textbf{77.3\%} & 53.9\% \\ \hline
    PIXOR-2 & 73.5\% & \textbf{72.1\%}
    \end{tabular}
    }
    \quad\\
    \quad\\
    \caption{Attack transferability between two PIXOR models with different seeds.}
    \label{table:pixor_transfer}
\end{minipage}
\quad
\begin{minipage}[b]{0.60\textwidth}
    \centering
    \resizebox{\textwidth}{!}{
    \begin{tabular}{ l | c | c | c | c}
    \backslashbox{Source}{Target} & PIXOR & PIXOR (d) & PointRCNN & PointPillar \\
    \hline
    PIXOR~\cite{pixor} & \textbf{77.3\%} & 66.0\% & 20.1\% & 8.2\% \\
    PIXOR (density)~\cite{pixor} & 66.4\% & \textbf{80.9\%} & 20.0\% & 7.7\% \\
    PointRCNN~\cite{pointrcnn} & 33.3\% & 33.7\% & \textbf{32.3\%} & 20.5\% \\
    PointPillar~\cite{pointpillar} & 54.9\% & 38.4\% & 28.4\% & \textbf{57.5\%}
    \end{tabular}
    }
    \label{tab:1}
    \vspace{-0.1in}
    \caption{Attack transferability among different detector models.}
    \label{table:transferability}
\end{minipage}
\vspace{-0.1in}
\end{table*}
  
\vspace{-0.05in}
\subsection{Datasets}
\label{sec:datasets}
\vspace{-0.05in}
We use the KITTI dataset~\cite{kitti} for training and evaluation of our attacks. KITTI
contains LiDAR point clouds and 3D bounding box labels for objects seen by the front camera of the
autonomous vehicle. For our experiments, we focus on the ``Car'' class only and consider each object
in a scene as a separate sample. Since our method relies on fitting meshes to point clouds, we
discard all samples with less than 10 points. This results in 6864 vehicles in the training set and
6544 vehicles in the validation set. We do not use the test set as labels are not publicly
available. For evaluation we only consider bounding boxes from a bird's eye view.


\subsection{Target LIDAR Detector Models}
\label{sec:targets}
In this work, we attack detector models that process point clouds exclusively without any auxiliary
inputs, since we only learn the shape of the adversarial mesh. We cover a variety of detector
architectures with different input representations of the point cloud. Specifically we consider the
following models:
\begin{itemize}[noitemsep]
\item \textbf{PIXOR}~\cite{pixor} is a detection network that processes input point
clouds into occupancy voxels and generates bounding boxes in a bird's eye view.

\item \textbf{PIXOR (density)} is a variant of PIXOR using density voxels as inputs. The value of
each voxel is calculated from bilinear interpolation of nearby points' distance to the voxel center:
$\sum_p (1 - |p_x - v_x|) (1 - |p_y - v_y|) (1 - |p_z - v_z|)$. The density variant allows us to compute
gradients to point clouds easier.

\item \textbf{PointRCNN}~\cite{pointrcnn} does not voxelize and instead processes the raw point cloud directly using a PointNet++ \cite{qi2017pointnet++} backbone. 
\item \textbf{PointPillar}~\cite{pointpillar} groups input points into discrete bins from BEV and uses PointNet~\cite{pointnet} to extract features for each pillar. 
\end{itemize}
\vspace{-0.05in}
Since we limit to the scope of learning the mesh shape only, we use the version of the above detectors that do not take LiDAR intensity as input.

\subsection{Experimental Setup}
\label{sec:setup}
\paragraph{Implementation details:} In our experiments, we initialize the adversarial mesh to be a unit
isotropic sphere with 162 vertices and 320 faces and scale it by $\mb{b} = (b_x, b_y, b_z) = \
$(0.7m, 0.7m, 0.5m). Maximum global offset is 0.1m on the $x,y$ direction and no offset is allowed
on the $z$ direction to prevent the mesh from moving into the vehicle or hovering in mid air. For
the Laplacian loss, we set $\lambda = 0.001$. During simulation, rays are sampled according to specs
of the Velodyne HDL-64E sensor used to generate the datasets.

For the gradient-based optimization in white box attacks, we use Adam with learning rate 0.005. For
the genetic algorithm in black box attacks, we initialize mutation std at 0.05, mutation probability
at 0.01, use a population size of 16, and average 100 queries to compute fitness. We decay the
mutation std and probability by a factor of 0.5 if the running fitness has not improved in 100
generations.

\subsection{Evaluation Metrics}
We consider the following two metrics for attack quality:
\begin{itemize}[leftmargin=*, noitemsep]
\vspace{-0.1in}
\item \textbf{Attack success rate:} Attack success rate measures the percentage at which the
target vehicle is successfully detected originally and but not detected after the attack. We
consider a vehicle successfully detected if the output IoU is greater than 0.7.
\item \textbf{Recall-IoU curve:}
Since attack success rate depends on the IoU threshold, we also plot the recall percentage at a
range of IoU threshold to get a more thorough measure of attack quality.
\end{itemize}


\vspace{-0.15in}
\begin{figure}[H]
\centering
\includegraphics[trim=0.2cm 0.2cm 0.3cm 0.3cm, clip,width=0.85\linewidth]{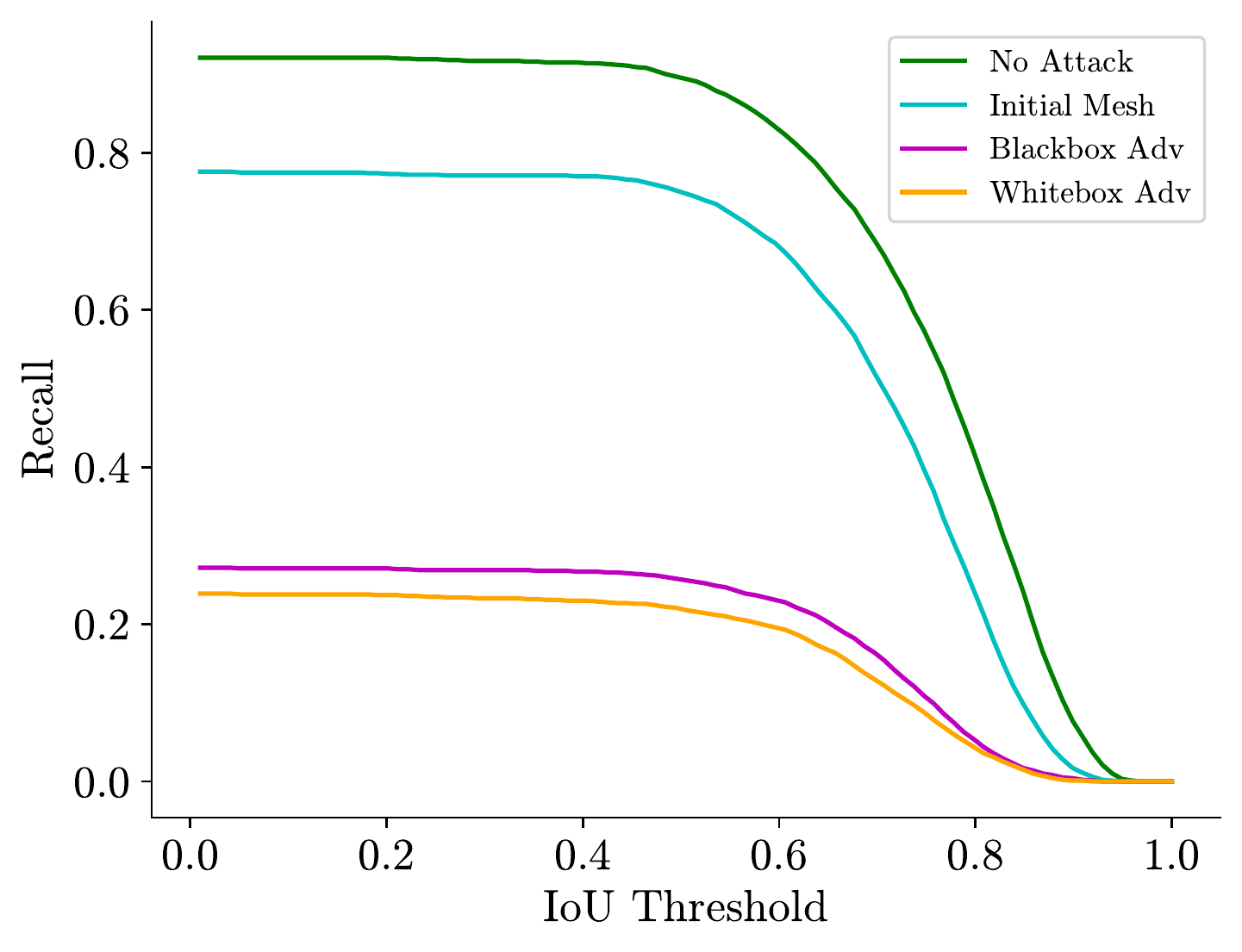}
\vspace{-0.1in}
\caption{Visualization of detection recall across range of IoUs for PIXOR with density voxels.
Difference between white box and black box attacks is very small. Initial mesh does not affect
detection as much as adversarial mesh.}
\label{fig:w_b_comp}
\vspace{-0.15in}
\end{figure}

\vspace{-0.05in}
\subsection{Results and Discussion}
\paragraph{Comparison of White Box and Black Box}
\vspace{-0.05in}
We conduct a white box and black box attack on a variant of PIXOR with density voxels. We visualize
the IoU-recall curve for both experiments in Figure~\ref{fig:w_b_comp}. 
and show that the attacks significantly drop the recall. When we use the initial icosphere mesh as a baseline, it has little
impact on detection even though it is of the same size as the adversary. We further compare our
black box attack against the white box alternative and show that they achieve similar performance.

\begin{figure*}[t]
\centering
\vspace{-0.15in}
\includegraphics[width=1.02\linewidth]{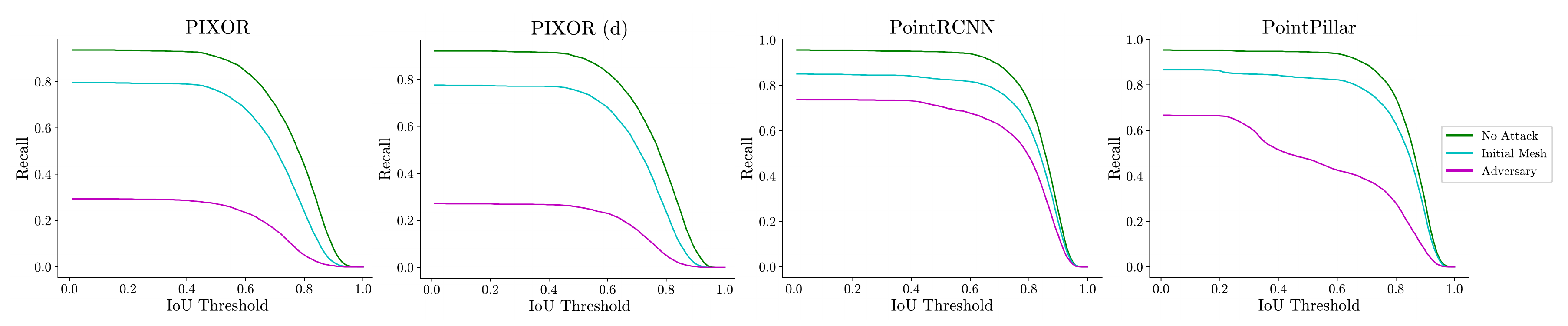}
\vspace{-0.15in}
\caption{IoU-Recall curve for black box attacks on different detector architectures.}
\label{fig:iou_recs}
\end{figure*}

\begin{figure*}[t]
\centering
\vspace{-0.1in}
\includegraphics[width=.94\linewidth]{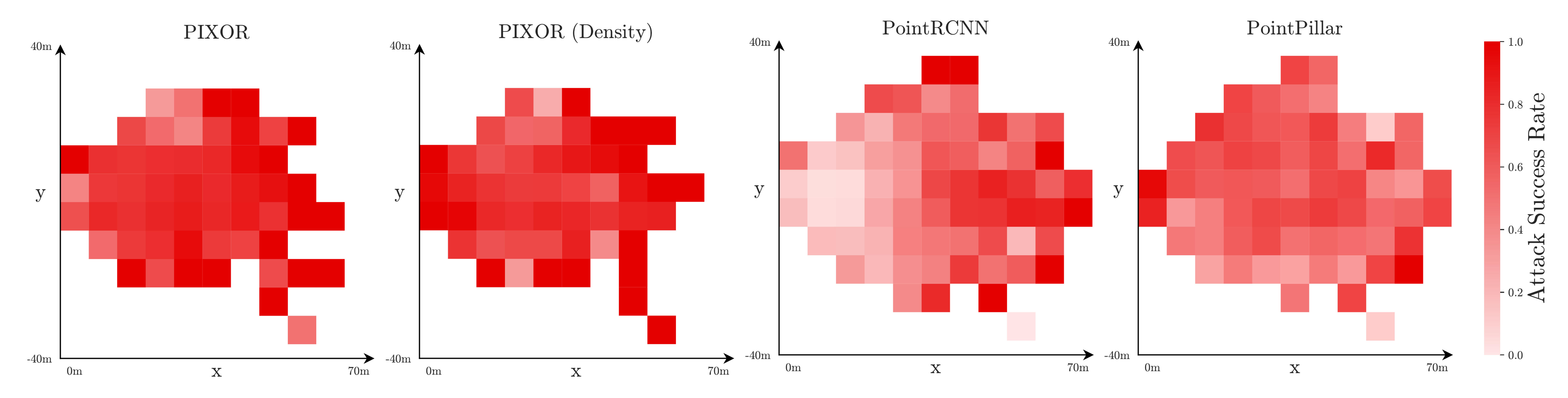}
\vspace{-0.2in}
\caption{Bird's eye view visualization of attack success rate at various locations in the scene using different detector models. 
}
\vspace{-0.15in}
\label{fig:heatmap}
\end{figure*}



    

\vspace{-0.1in}
\paragraph{Transferability Across Identical Architectures}
\vspace{-0.05in}
We investigate the transferability of adversarial examples across similar models. To this end, we
train two variations of the original PIXOR model using different seeds and learn adversarial meshes
for each model separately. We then evaluate transferability between the pair of models using attack
success rate as the metric. Results are summarized in Table~\ref{table:pixor_transfer} and there is
a high degree of transferability between models with identical architecture. This allows strong transfer attacks with only knowledge of model architecture.~\cite{transfer}.
\newcommand{\makerow}[9]{
    \textbf{#1} & 
    \centering \raisebox{-0.4\totalheight}{\includegraphics[width=.65\linewidth]{#4}}
    & \centering #2\%  & \centering #6 \% &
    \centering \raisebox{-0.4\totalheight}{\includegraphics[trim=0cm 0cm 0cm #9cm, clip, width=.65\linewidth]{#5}} 
    & \centering \textbf{#3\%}  & \centering #7 \% & #8\\
}
\begin{table*}[t]
    \centering
    \vspace{-0.05in}
    \resizebox{\textwidth}{!}{
    \begin{tabular}{ c | p{2.0cm} p{1.2cm} p{2cm} | p{2.0cm} p{1.2cm} p{2cm} | c }
    Object & \centering Initial & \centering Success & Classification &
    \centering Adversarial & Success & Classification & Dimensions \\
    \hline
    \makerow{Couch}{23.3}{68.6}{common/base_couch.png}{common/adv_couch.png}{93.1}{93.8}{1.69m x 0.85m x 0.94m}{-0.5} 
    \makerow{Canoe}{26.9}{59.5}{common/base_canoe.png}{common/adv_canoe.png}{99.6}{99.9}{3.51m x 0.81m x 0.68m}{0}
    \makerow{Table}{18.9}{48.0}{common/base_table.png}{common/adv_table.png}{99.8}{99.7}{1.57m x 0.83m x 0.86m}{0}
    
    \makerow{Cabinet}{20.7}{54.1}{common/base_cabinet.png}{common/adv_cabinet.png}{93.4}{94.2}{1.29m x 0.91m x 0.76m}{0}
    \makerow{Chair}{14.1}{23.3}{common/base_chair.png}{common/adv_chair.png}{99.9}{99.9}{1.42m x 0.64m x 0.71m}{0}
    \makerow{Bike}{19.6}{32.4}{common/base_bike.png}{common/adv_bike.png}{94.4}{92.3}{1.70m x 0.76m x 1.08m}{0}

    \end{tabular}
    }
    \label{tab:1}
    \vspace{-0.1in}
    \caption{Adversaries resembling common objects that could appear on the rooftop. Attack success rates on the initial and adversarial configurations are shown. A 
    ShapeNet classifier stably recognizes our adversaries as the correct object class.}
    \label{table:common_obj}
\end{table*}

\begin{table*}[t]
    \vspace{-0.05in}
    \centering
    \begin{tabular}{l | c | c | c | c | c | c | c || c }
     & Arbitrary & Couch & Canoe & Table & Cabinet & Chair & Bike & AP \\ \hline
    Original & 77.3\% & 68.6\% & 59.5\% & 48.0\% & 54.1\% & 23.3\% & 32.4\% & 74.37\\ 
    Augmentation & 14.4\% & 12.4\% & 19.6\% & 11.5\% & 7.3\% & 6.6\% & 14.0\% & \textbf{74.92}\\ 
    Adv Train & \tb{10.8\%} & \tb{5.6\%} & \tb{18.7\%} & \tb{4.9\%} & \tb{5.3\%} & \tb{6.3\%} & \tb{11.4\%} & 73.97
    \end{tabular}
    \vspace{-0.1in}
    \caption{Attack success rates before and after applying defense training on PIXOR. In the final column, we evaluate the models on the standard KITTI validation set and show the average precision (AP) at 0.7 IoU. 
    }
    \label{table:defense}
    \vspace{-0.15in}
\end{table*}

\vspace{-0.15in}
\paragraph{Transferability Across Input Representations}
\vspace{-0.05in}
In this section we attack four models described in Section~\ref{sec:targets}, and use only black box
attacks due to non-differentiable layers in some models. We provide IoU-Recall curves in
Figure~\ref{fig:iou_recs}. Again, transferability between the models is considered and results are
shown in Table~\ref{table:transferability}.

First, our attack is able to hide objects with high probability on the PIXOR models and PointPillar
but is significantly weaker on PointRCNN. We hypothesize that this is because PointRCNN treats every
point as a bounding box anchor. Therefore, vehicles close to the sensor register significantly more
LiDAR points and proposals, making it extremely difficult to suppress all proposals. We verify this
hypothesis by visualizing the attack success rate at various locations across the scene in
Figure~\ref{fig:heatmap}. The success rate on attacking PointRCNN is close to 0 near the LiDAR sensor but
grows substantially higher as the distance to the sensor increases and the number of points decreases.

In terms of transferability, the occupancy and density variants of PIXOR can share a significant
portion of adversarial examples. The attacks generated from PointPillar and PointRCNN can also be
used to attack PIXOR, but not vice versa. This suggests that additional layers of point-level
reasoning before aggregating on the $z$-dimension probably make the model more robust to
rooftop objects.

In addition, two variations of PIXOR have different vulnerable regions even though they share the
same backbone. Specifically, we note that vehicles close to the LiDAR sensor are the easiest
targets when using a density voxel input representation. In contrast, vehicles closeby are the most
robust to attacks when using occupancy voxels. We speculate that this is an effect of the input
precision. For density voxels, the number of points is significantly higher near the LiDAR sensor,
contributing to higher input precision, whereas occupancy voxels can only show a binary indicator
whether a voxel contains a point or not. 

\begin{figure}[h!]
\vspace{-0.05in}
\centering
\includegraphics[trim=0.2cm 0.3cm 0.3cm 0.3cm, clip, width=.8\linewidth]{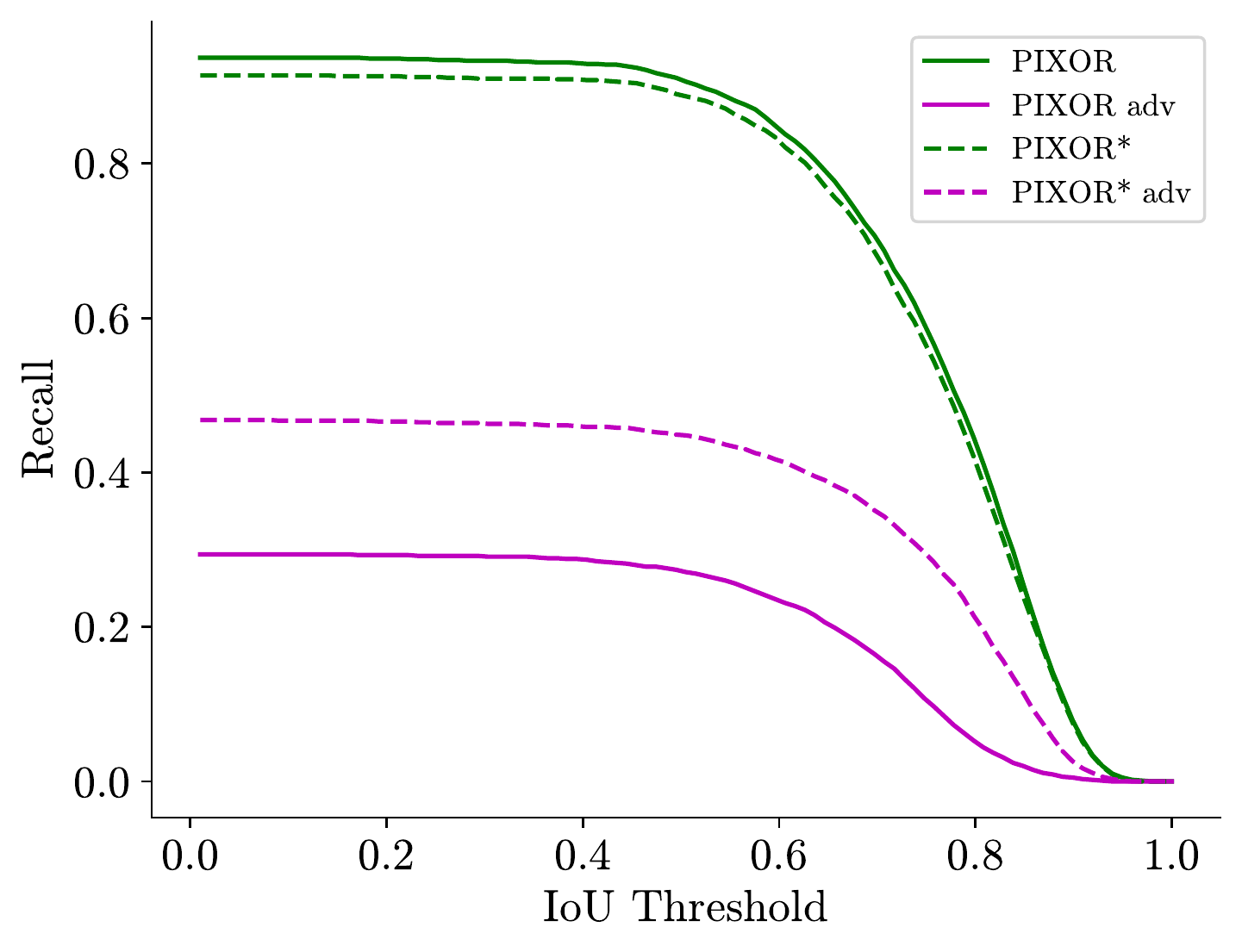}
\vspace{-0.1in}
\caption{We perform our attack on two similar models. PIXOR converts the input point cloud to occupancy voxels and PIXOR* separates points into columns and extracts features with a pointnet backbone. Although the models have the same backbone architecture, PIXOR* is significantly more robust due to the input representation. 
}
\label{fig:rep_comp}
\vspace{-0.1in}
\end{figure}

Based on the above observations, we conclude that the choice of input representation and detection
scheme may have significant implications on robustness to adversarial examples. For a more concrete comparison, we consider a variant of PIXOR using PointPillar's pillar representation instead of voxelization and keep the backbone architecture identical. We compare this variant, PIXOR* against PIXOR and show the results in Figure \ref{fig:rep_comp}. Here, we can see that with even with identical backbones and training routines, PIXOR* is significantly more robust to our attack purely due to a different input representation.   

\vspace{-0.1in}
\paragraph{Common Objects}
\vspace{-0.1in}
In this section, to make attacks more realistic, we learn adversaries that resemble common objects
that may appear on top of a vehicle in the real world. Instead of deforming an icosphere, we
initialize from a common object mesh and deform the vertices while constraining the maximum
perturbation distances. We choose couch, chair, table, bike, and canoe as six common object classes,
and we take the object meshes from ShapeNet~\cite{shapenet}. We apply uniform mesh re-sampling in
meshlab~\cite{meshlab} to reduce the number of faces and produce regular geometry prior to
deformation. In these experiments we limit the maximum vertex perturbation to 0.03m so that the
adversary will resemble the common object, and limit translation to 0.1m, and allow free rotation.
In Table~\ref{table:common_obj}, we present the visualizations, results, and dimensions of the
common objects. Moreover, the identity of the adversarial objects are unambiguous to a human, and we
also verify that a PointNet~\cite{qi2017pointnet++} classifier trained on ShapeNet~\cite{shapenet} is also able to correctly
classify our perturbed objects. This confirms the possibility that the placement of common objects
can also hurt LiDAR detectors.

\vspace{-0.1in}
\paragraph{Adversarial Defense}
\vspace{-0.08in}
We employ our proposed defense methods by retraining a PIXOR model with random data augmentation and adversarial training. 
To generate random meshes for data augmentation, we uniformly sample $N$ from $[50, 200]$, $M$ from $[0, 300]$, and we sample vertices from Gaussian $\mathcal{N}(0, 0.5)$.
If all tetrahedrons are removed by decimation, the sampling process restarts.
During training, for every scene we sample one vehicle at random for data augmentation.
We only augment vehicles with at least 10 points in the scene, otherwise it is too difficult to fit a mesh for roof approximation. During adversarial training, we set $k=30$ and alternate between updating the mesh and the model. 

For evaluation, we re-train an adversarial mesh on the defended model and observe that the attack
success rate is reduced significantly, as shown in Table~\ref{table:defense}. In addition, we also
launch attacks with common objects on the defended model and observe similar findings. 
Furthermore, for the standard detection task, our defended models achieve similar or better performance when evaluated on the KITTI validation set.
Nevertheless, the defense is not yet perfect since there is still 5-20\% attack success rate remaining. With more computational resources, full adversarial training could possibly close this gap. 

\vspace{-0.05in}
\section{Conclusion}
\vspace{-0.05in}
We propose a robust, universal, and physical realizable adversarial example capable of hiding
vehicles from LiDAR detectors. An attacker can 3D print the mesh and place it on any vehicle to make
it ``invisible'' without prior knowledge of the scene. The attack will consistently cause target
vehicles to disappear, severely impeding downstream tasks in autonomous driving systems. Even without
any malicious intent, we show that problematic shapes can coincidentally appear with common objects
such as a sofa. We further show that training with data augmentation using random meshes can
significantly improve the robustness, but unfortunately still not 100\% secure against our attack.
By demonstrating the vulnerability of LiDAR perception against universal 3D adversarial objects, we
emphasize the need for more robust models in safety-critical robotics applications like
self-driving.

{
\small
\bibliographystyle{ieee_fullname}
\bibliography{bib}
}

\end{document}